\documentclass[11pt]{article}
\usepackage{coling2020}
\usepackage{times}
\usepackage{url}
\usepackage{latexsym}
\usepackage{graphicx}
\usepackage{verbatim}
\usepackage{hyperref}
\usepackage{subfig}
\usepackage{float}
\usepackage{xcolor}
\usepackage{amsmath}
\usepackage{booktabs}
\usepackage{comment}
\usepackage{textcomp}
\usepackage{xspace}
\usepackage[hide]{boxnotes}

\newcommand{\eat}[1]{} 
\newcommand{\laks}[1]{\textcolor{purple}{#1}} 

\colingfinalcopy 

\newcommand{\gan}[1]{\noindent{\textcolor{black}{\{{$_{gan}$} \bf   #1\}}}}

\iftrue 
  \renewcommand{\laks}[1]{#1}
  \renewcommand\gan[1]{\xspace}
\fi

\title{Automatic Detection of Machine Generated Text: A Critical Survey}

\author{Ganesh Jawahar, Muhammad Abdul-Mageed, Laks V.S. Lakshmanan \\ 
University of British Columbia, Vancouver, Canada \\ 
{\tt ganeshjwhr@gmail.com}, {\tt muhammad.mageed@ubc.ca}, {\tt laks@cs.ubc.ca}}

\date{}

\begin{document}
\maketitle
\begin{abstract}
Text generative models (TGMs) excel in producing text that matches the style of human language \laks{reasonably well.} Such TGMs can be misused by adversaries, e.g., by automatically generating fake news and fake product reviews that can look authentic and fool humans. Detectors that can distinguish text generated by TGM from human written text play a vital role in  mitigating such misuse of TGMs. Recently, there has been a flurry of works from both natural language processing (NLP) and machine learning (ML) communities to build accurate detectors for English. Despite the importance of this problem, there is currently no work that surveys this fast-growing literature and introduces newcomers to \laks{important research} challenges. In this work, we \laks{fill this void} by providing a critical survey and review \laks{of this literature} to facilitate \laks{a comprehensive} understanding of this problem. We conduct an in-depth error analysis of the state-of-the-art detector and discuss research directions to guide future work in this exciting area.  
\end{abstract}

\section{Introduction}
Current state-of-the-art text generative models (TGMs) excel in producing text that \laks{approaches}  the style of human language, especially in terms of grammaticality, fluency, coherency, and usage of real world knowledge~\cite{radford2019language,zellers_neurips19,keskar_arxiv19,bakhtin_arxiv20,brown2020language}. TGMs are useful in a wide variety of applications, including story generation~\cite{fan_acl18}, conversational response generation~\cite{zhang_arxiv19}, code auto-completion~\cite{solaiman_arxiv19}, and radiology report generation~\cite{liu_mlhc19}.
However, TGMs can also be misused for 
fake news generation~\cite{zellers_neurips19,brown2020language,uchendu_emnlp20}, fake product reviews generation~\cite{adelani_aina20}, and spamming/phishing.~\cite{weiss_ts19}.~\footnote{\url{https://www.businessinsider.com/fake-ai-generated-gpt3-blog-hacker-news-2020-8}} 
Thus, it is important to build tools that can minimize the threats posed by \laks{the} misuse of TGMs.

The commonly used approach to combat the threats posed by the misuse of TGMs is to formulate the problem of distinguishing text generated by TGMs and human written text as a classification task. 
The \laks{classifier, henceforth called} \textit{detector}, can be used to automatically remove machine generated text from online platforms such as social media, e-commerce, email clients, and government forums, \laks{when the intention of the TGM generated text is abuse}. An ideal detector should be: (i) \textit{accurate}, that is, good accuracy with a good trade-off for false positives and false negatives depending on the online platform (email client, social media) on which TGM is applied~\cite{solaiman_arxiv19}; (ii) \textit{data-efficient}, that is, needs as few examples as possible from the TGM used by the attacker~\cite{zellers_neurips19}; (iii) \textit{generalizable}, that is, detects text generated by different modeling choices of the TGM used by the attacker such as model architecture, TGM training data, TGM conditioning prompt length, model size, and text decoding method~\cite{solaiman_arxiv19,bakhtin_arxiv20,uchendu_emnlp20}; and (iv) \textit{interpretable}, that is, detector decisions need to be understandable to humans~\cite{gehrmann_acl19}; and (v) \textit{robust}, that is, detector can handle adversarial examples~\cite{wolff2020attacking}. 
Given the importance of this problem, there has been a flurry of \laks{research} recently from both NLP and ML communities \laks{on} building useful detectors. However, there is currently no work that provides a literature review of existing detection works and highlight important research challenges. 

In this paper, we present a critical literature review of the existing detection research for English to aid understanding of this important area. We organize the survey to guide the reader seamlessly through a number of important aspects, as follows:  First, we establish the background for the detection task, \laks{which} includes TGMs, decoding methods for text generation, and social impacts of TGMs (\textsection\ref{sec:bg}). Second, we present various aspects of large-scale TGMs such as model architecture,
training cost, and controllability 
(\textsection\ref{sec:threat}). Third, we present and discuss the various existing detectors in terms of their underlying methods (\textsection\ref{sec:det}). Fourth, we provide a linguistically and computationally motivated analysis of key issues of the state-of-the-art detector 
(\textsection\ref{sec:roberta_issues}). Fifth, we discuss interesting future research directions that can help in building useful detectors (\textsection\ref{sec:future}). Our main contributions are three-fold:
\begin{itemize}
  \item We provide the first survey on the important, burgeoning area of detection of machine generated text from human written text.
  \item We develop an error analysis of current state-of-the-art detector, guided and illustrated by machine generated texts, to shed light on the limitations of existing detection work. 
  \note[Laks]{This bullet is confusing: error analysis of detectors guided by machine generated text? What would thta mean?}\gan{We manually look at the false positive reviews (machine generated text incorrectly classified as human text) of the detector. We come up with error categories thanks (guide) to these machine generation texts, some of which we put up (illustrate) as examples. Should I re-phrase?}
  \item Motivated by our analysis and existing challenges, we propose a rich and diverse set of research directions to guide future work in this exciting area.
\end{itemize}

\section{Background}
\label{sec:bg}
Here, we provide the background for the problem of detecting machine generated text from human written text. Specifically, we introduce key concepts in training a TGM, generating text from a TGM, and social implications of using TGMs in practice. Existing detection datasets are discussed in Appendix.

\subsection{Training TGM}
TGM is typically a neural language model (NLM) trained to model the probability of a token given the previous tokens in a text sequence, i.e.,  $p_\theta(x_t|x_1,\dots,x_i,\dots,x_{t-1})$, with tokens coming from a vocabulary, $x_i \in \mathcal{V}$.\note[Laks]{Above, wouldn't you want to say $x_i \in \mathcal{V}$ rather than $x_t \in \mathcal{V}$? \\ Also, you should justify why the survey focuses on neural models with traditional LM objectives. E.g., are they the SOTA? By fer the best? Something else?}\gan{They are the best. Added.} If $\mathbf{x} = (x_1,\dots,x_{|\mathbf{x}|})$ represents the text sequence, $p_{\theta}$ typically takes the form $p_{\theta}(\mathbf{x}) = \Pi_{t=1}^{|\mathbf{x}|} p_{\theta}(x_t|x_1,\dots,x_{t-1})$. \note[Laks]{Do previous works tend to make any assumptions such as conditional independence? You should state those.}\gan{Deep learning based language models do not make conditional independence assumption.}
If $p_{*}(\mathbf{x})$ denotes the reference distribution and $\mathcal{D}$ denotes a finite set of text sequences from $p_*$, TGM estimates parameters $\theta$ by minimizing the following objective function:
\begin{equation}\label{eq:orig-obj} 
\mathcal{L}(p_\theta, \mathcal{D}) = - \sum_{j=1}^{|\mathcal{D}|} \sum_{t=1}^{|\mathbf{x}^{(j)}|} \log p_\theta(x_t^{(j)} | x_1^{(j)},\dots,x_i^{(j)},\dots,x_{t-1}^{(j)}).
\end{equation}
Notice that TGM can be a non-neural model (e.g., n-gram LM) and based on nontraditional LM objective (e.g., masked language modeling~\cite{devlin_naacl19,song2019mass}). In this survey, we focus primarily on TGMs for English that are neural and based on traditional LM objective, as they are successful in generating coherent paragraphs of English text.


\subsection{Generating text from TGM}
\label{sec:related_gen_text_tgm}
Given a sub-sequence (\textbf{prefix}), $\mathbf{x}_{1:k} \sim p_*$, the task of generating text from TGM is to use $p_\theta$ to conditionally decode a \textbf{continuation}, $\hat{\mathbf{x}}_{k+1:N} \sim p_\theta(.|\mathbf{x}_{1:k})$ such that the resulting \textbf{completion} ($x_1,\dots,x_k,\hat{x}_{k+1},\dots,\hat{x}_N$) resembles a sample from $p_*$~\cite{Welleck2020Neural}.
In a news article generation task, the prefix can be headlines and the continuation can be the body of the news article. In a story generation task, the prefix can be beginning of a story and the continuation can be rest of the story. 
Since the computation of the optimal continuation ($\hat{\mathbf{x}}_{k+1:N}$) is not tractable with time complexity of $O((N-k)^{|\mathcal{V}|})$, approximate \textit{deterministic} or \textit{stochastic} decoding methods are utilized to generate continuations.

\noindent\textbf{\underline{Deterministic methods}}: In deterministic methods, the continuation is fully determined by the TGM parameters and prefix. The two most commonly used deterministic decoding methods are greedy search and beam search. \textit{Greedy search} works by selecting the highest probability token at each time step: $x_t = \arg \max p_\theta(x_t|x_1,\dots,x_{t-1})$ with time complexity of $O((N-k)|\mathcal{V}|)$. On the other hand, \textit{beam search} maintains a fixed-size ($b$) set of partially decoded sequences, called hypotheses. At each time step, beam search creates new hypotheses by appending each token in the vocabulary to each existing hypothesis, scoring the resulting sequences using $p_*$ with time complexity of $O((N-k)b|\mathcal{V}|)$. In practice, these deterministic decoding methods depend highly on the underlying model probabilities and suffer from producing degenerate continuation, i.e., generic text often with repetitive tokens~\cite{holtzman_iclr20}. Recently, Welleck et al.,~\shortcite{Welleck2020Neural} show that the degeneracy issues with beam search can be alleviated by training a TGM with \laks{the} original TGM objective (Eq.~(\ref{eq:orig-obj})) augmented with \laks{an} unlikelihood objective that assigns lower probabilities to unlikely generations.\note[Laks]{You may also want to comment on the complexity of these methods, esp. given that the otimal is intractable. Beam seacrh sounds expensive compared to greedy search.}\gan{Added.}

\noindent\textbf{\underline{Stochastic methods}}: Stochastic decoding methods work by sampling from a model-dependent distribution at each time step, $x_t \sim q(x_t|x_1,\dots,x_{t-1}, p_\theta)$. In unrestricted sampling (also known as \textit{pure sampling}), the chance of sampling a low-confidence token from the unreliable tail distribution is very high, leading to text that can be unrelated to prefix. To reduce the chance of sampling a low-confidence token, sampling is limited to a subset of the vocabulary $\mathcal{W} \subset \mathcal{V}$ at each time step. Let $\mathcal{Z} = \sum_{x\in\mathcal{W}} p_\theta(x|x_1,\dots,x_{t-1})$. If $x_t \in \mathcal{W}$, $q(x_t|x_1,\dots,x_{t-1},p_\theta)$ is set as $p_\theta(x_t|x_1,\dots,x_{t-1})/\mathcal{Z}$, otherwise \laks{set as}  $0$.
The two most effective stochastic decoding methods are \textit{top-k sampling}~\cite{fan_acl18} and \textit{top-p} (or \textit{nucleus}) \textit{sampling}~\cite{holtzman_iclr20}. The top-k sampler limits sampling to the $k$ most-probable tokens, that is, $\mathcal{W}$ is the size $k$ subset of $\mathcal{V}$ that maximizes $\sum_{x\in\mathcal{W}} p_\theta(x|x_1,\dots,x_{t-1})$. The top-k sampler uses a constant value of $k$, which can be sub-optimal in different contexts, that is, generated text is limited to a subset of natural language distribution. For example, generic contexts (e.g., predicting noun) might require larger value of $k$,  while other contexts \laks{(e.g., predicting prepositions)} might require smaller value of $k$ so that only useful candidate tokens are considered. The nucleus sampler overcomes the burden of considering only a fixed number of tokens 
by limiting sampling to the smallest set of tokens with total mass above a threshold $p \in [0, 1]$, \laks{i.e.,}  $\mathcal{W}$ is the smallest subset with $\sum_{x\in\mathcal{W}} p_\theta(x|x_1,\dots,x_{t-1}) >= p$. Thus, the number of candidate tokens considered varies  dynamically depending on the context, and the resulting text is reasonably natural with less repetitions. 
Recently, Massarelli et al.,~\shortcite{massarelli2020decoding} show that top-k and top-p sampler tend to generate more nonfactual sentences, as corroborated by Wikipedia.
 
\subsection{Social impacts of TGMs}

\noindent\textbf{\underline{Bias}}: Unsurprisingly, a TGM can capture and amplify the societal biases (over-generalized beliefs about a particular group of people, e.g., \textit{Group X are bad drivers}) present in the training data~\cite{sun-etal-2019-mitigating,nadeem2020stereoset}. Solaiman et al.,~\shortcite{solaiman_arxiv19} and Brown et al.,~\shortcite{brown2020language} show that TGMs reflect gender bias (e.g., favoring males over females), racial bias (e.g., favoring white over black people), and religious bias (e.g., favoring Christians over Muslims).\note[Laks]{Did you mean Islamists here or Muslims?}\gan{``Islam religion" verbatim from GPT-3 paper.} Although TGMs can be used as a tool to study how patterns in the training data can translate to these unintended biases in the model outputs~\cite{solaiman_arxiv19}, the biases can cause harm to the people in relevant groups in many ways~\cite{bias_trouble}. 

\noindent\textbf{\underline{Beneficial usage}}: 
TGMs are used to create task-specific systems, such as question answering, reading comprehension, natural language inference, and machine translation~\cite{radford2019language,brown2020language}. TGMs can also be used to generate text that \laks{approximately} matches the style of human language, which \laks{benefits} applications such as story generation~\cite{fan_acl18}, conversational response generation~\cite{zhang_arxiv19}, code auto-completion~\cite{tab_nine}, and radiology report generation~\cite{liu_mlhc19}. 

\noindent\textbf{\underline{Malicious usage}}: 
TGMs can have unfortunate uses by (even low-skilled) adversaries for malicious purposes, such as fake news generation~\cite{zellers_neurips19,brown2020language,uchendu_emnlp20}, fake product reviews generation~\cite{adelani_aina20}, and spamming/phishing~\cite{weiss_ts19}. Humans can spot fake news articles~\cite{brown2020language}, fake product reviews~\cite{adelani_aina20}, and fake comments~\cite{weiss_ts19} generated by TGM only at chance level. 
To combat the threats posed by such adversaries, accurate models that can identify text generated by TGM from human written text need to be built. Such a model can have benevolent uses such as moderating content in vulnerable platforms including social media, email clients, government websites, and e-commerce websites.

\begin{table}[]
\footnotesize
\centering
\begin{tabular}{p{0.6in}|p{1.4in}|p{1.1in}|p{0.95in}|p{0.6in}|p{0.5in}}  \hline
\textbf{TGM}   & \textbf{training text sequence} ($\mathbf{x}$) (data size / params)  & \textbf{prefix} ($\mathbf{x}_{1:k}$) & \textbf{continuation} ($\hat{\mathbf{x}}_{k+1:N}$)  & \textbf{decoding method} & \textbf{threats discussed} \\  \hline 
GPT-2 \cite{radford2019language} & fragments from WebText (collection of internet articles) (40GB / 1.5B) & starting of an article (e.g., few lines about a research finding) & rest of the article (e.g., rest of the research finding) & top-k    & NA       \\ \hline
GROVER \cite{zellers_neurips19} & news article along with their meta-information from RealNews (120GB / 1.5B) & meta-information/body of a news article (e.g., headline, author) & missing meta-information/body in the prefix & top-p & trustworthy fake news \\ \hline
CTRL \cite{keskar_arxiv19} & control code (e.g., URL) followed by text (e.g., news article) from several domains (140GB / 1.6B) & control code (e.g., URL) with optionally some strings of text & article corresponding to the control code & greedy search with repetition penalty & NA \\  \hline
Adelani et al.,~\shortcite{adelani_aina20} & product reviews (fine-tuning GPT-2) (20GB / 0.1B) & product review (human written) & product review (machine) & top-k & fake prod. reviews \\ \hline
Dathathri et al.,~\shortcite{dathathri_iclr20} & no training and no fine-tuning & beginning of a story or general articles & rest of the story or article & top-k & NA \\ \hline
GPT-3 \cite{brown2020language} & fragments from CommonCrawl (570GB / 175B) & three previous news articles and title of a proposed article & body of the proposed article & top-p & fake news \\
\bottomrule
\end{tabular}
\caption{Summary of the characteristics of TGMs that can act as threat models. The last column corresponds to the threats discussed in the original paper.}
\label{tab:threat_models}
\end{table}

\section{Text generative models}
\label{sec:threat}
In this section, we will discuss various aspects of large-scale TGMs. These TGMs act as threat models since they can be misused by a low-skilled adversary, \laks{e.g.,}  by generating fake news and fake product reviews. Table~\ref{tab:threat_models} displays the summary of key characteristics of these TGMs along with the threats they \laks{pose} (according to the original papers).

\subsection{Model architecture, training data, training cost}
\noindent\textbf{\underline{Model architecture}}: The model architecture underlying all the state-of-the-art TGMs is the transformer~\cite{vaswani_neurips17}. Compared to recurrent neural networks (RNNs)~\cite{ELMAN1990179}, the transformer model does not have a bias to recent tokens and can learn long-range dependency information. 
The generation from TGMs such as GPT-2 \laks{which are} based on transformer architecture tends to be grammatically correct, coherent, and uses world knowledge~\cite{radford2019language}.~\footnote{Text generated by RNN can be more easily detected~\cite{fagni2020tweepfake}, as such text is usually less grammatically correct and less coherent (based on our manual observations).}

\noindent\textbf{\underline{Training data}}: TGMs such as GPT-2, CTRL~\cite{keskar_arxiv19}, and  GPT-3~\cite{brown2020language} have billions of parameters. They are generally trained using the language modeling objective
on large amounts of raw text from a diverse set of sources (like Wikipedia, Reddit, and news sources). As an exception, GROVER~\cite{zellers_neurips19} is trained on millions of news article only. 
Such trained TGMs can also be fine-tuned on a domain-specific corpus for the LM task to generate text that matches the respective domain reasonably. For example, Adelani et al.,~\shortcite{adelani_aina20} fine-tune the GPT-2 model on the specific domain of product reviews to generate fake reviews, which mimics the style of a human review. 

\noindent\textbf{\underline{Training cost}}: Training TGMs with billions of parameters on millions of documents requires a huge computational budget~\cite{zellers_neurips19}, high energy cost~\cite{strubell-etal-2019-energy},  and \laks{long} training time~\cite{brown2020language}. Unfortunately, it is not yet a standard practice to report financial (vs. energy vs. computational) budget in every research publication. This makes it hard for us to perform TGM training feasibility studies. One exception is the work done by Zellers et al.,~\shortcite{zellers_neurips19}, where they explicitly mention that their proposed TGM model, GROVER, took two weeks of training with a cost of \$25K (including the cost of data collection). We note that even though this may be an expensive budget, it is by no means outside the reach of even low-resource organizations, let alone nation states. \eat{As such,}\laks{The implication is that} various entities of variable sizes and resource capabilities can practically deploy models for spreading disinformation using TGMs. 

\subsection{Controllable generation}
\label{sec:controllable}
Controllable TGMs possess the ability to control the aspects of the generation such as topic and sentiment of the article. GPT-2~\cite{radford2019language} and GPT-3~\cite{brown2020language} assume the prefix to be \laks{any} natural language text, which might be too \laks{coarse} in controlling the generation in an explicit fashion. Researchers have devised two ways to design a controllable TGM, which we now introduce. 

\noindent\textbf{\underline{Training with control tokens}}: The first way is to leverage meta-information about the article such as its author, date of creation, source domain and prepend \laks{this information} as additional token(s) to the input sequence, before training the TGM. These tokens act as additional context for the article, allowing the TGM to learn the relation between the meta-information and the original article. Once trained, the TGM model can be controlled by prompting with the meta-information of users' interest. The first controllable TGM \laks{proposed} is the GROVER model, \laks{which} can generate a news article given the meta-information of the news article (such as headline, author, and date). \laks{The} GROVER model can create trustworthy fake news that is harder for humans to identify than human written fake news and can thus pose a significant threat.\note[Laks]{Just a clarification. You are saying GROVER can thus pose a significant threat, right?}\gan{Yes, exactly. Added.} Similar to the GROVER model, the CTRL model provides explicit control of particular aspects of the generated text by exploiting naturally occurring control codes (e.g., the URL for a news article) to condition the text (e.g., news article body). These control codes govern style (e.g., sports vs. politics, FOX sports vs. CNN sports), content (e.g., Wikipedia vs. books), and task-specific behavior (e.g., question answering vs. machine translation). 
\note[Laks]{Sports vs. politics also concern different *domains*. It is just that the styles followed in articles in those domains can be quite different. For contrast, consider NYT News vs. Fox News or some fringe outlet news. Even withinm one domain, say politics, the *style* of articles in these outlets are very different. You may want to clarify if this sort of style can be controlled by the above approach.}\gan{Yes, it can be controlled at this level. Added another example for style. Is it ok?}

\noindent\textbf{\underline{Control using attribute classifier}}: The second and the most recent way to design a controllable TGM is to combine a pretrained TGM like GPT-2 with one or more attribute classifiers (e.g., sentiment classifier) that guide text generation~\cite{dathathri_iclr20}. 
The attribute models measure the extent to which the desired attribute is encoded in a piece of text. At each timestep, GPT-2 updates its latent representations based on gradients from the attribute model for the text generated so far \laks{so as} to increase the likelihood of the generated text having the desired attribute. The updated latents are used to compute a new next token distribution from which a token to be generated is sampled. 
The interesting property of 
this method is that the TGM model need not be retrained
(unlike Adelani et al., \shortcite{adelani_aina20} work that need retraining of the GPT-2 model), thereby avoiding the significant cost of retraining.

\section{Detectors}
\label{sec:det}
In this section, we discuss various detectors for identifying machine generated text from human written text. To aid understanding of the literature, we organize the detectors according to the underlying methods on which they are based.

\subsection{Classifiers trained from scratch}
\label{sec:classi_from_scratch}
\noindent\textbf{\underline{Bag-of-words classifier}}: Some detectors employ classical machine learning methods such as logistic regression to train a model from scratch to discriminate between text generated by TGM and human written text. Solaiman et al.,~\shortcite{solaiman_arxiv19} use a simple baseline model that represents a document with tf-idf vector (unigrams and bigrams) on top of a logistic regression model to distinguish WebText articles (online web pages) \laks{from text generated using GPT-2 models.} 
They study different sizes of GPT-2 models that vary in terms of number of parameters (117M, 345M, 762M, 1542M) and different sampling techniques (pure sampling, top-k sampling, and top-p sampling). 
They observe that generations from the larger GPT-2 models are difficult to detect compared to that of the smaller models, which indicates that the larger the TGM, the closer the style of the generated text with that of human written text. Top-k samples are easier to detect while nucleus samples are harder to detect. 
This result stems from the fact that top-k sampler typically over-generates common words, leaving statistical anomalies that are easily spotted by the detector~\cite{ippolito_arxiv19}.  
Additionally, Solaiman et al.,~\shortcite{solaiman_arxiv19} fine-tune the GPT-2 model on Amazon product reviews and show that the text generated by fine-tuned GPT-2 model is harder to detect as fine-tuned domain specific TGMs are more human-like than general purpose TGM (i.e., the original GPT-2 model). 

\noindent\textbf{\underline{Detecting machine configuration}}: Tay et al.,~\shortcite{tay_arxiv20} study the extent to which different modeling choices (decoding method, TGM model size, prompt length) leave artifacts (detectable signatures that arise from modeling choices) in the generated text. They propose the task of identifying the TGM modeling choice given the text generated by TGM. They show that a classifier can be trained to predict the modeling choice well beyond the chance level, which ascertains that text generated by TGM may be more sensitive to TGM modeling choices than previously thought. 
They also find that the proposed detection task of identifying text generated by different TGM modeling choices is less harder than the task of identifying text generated by TGM from human written text along with different TGM modeling choices. They show that word order does not matter much as a bag-of-words detector performs very similar to detectors based on complex encoder (e.g., transformer). This result is consistent with the recent work done by Uchendu et al.,~\shortcite{uchendu_emnlp20}, which shows that simple models (traditional ML models trained on psychological features and simple neural network architectures) perform well in three settings: (i) classify if two given articles are generated by the same TGM; (ii) classify if a given article is written by a human or a TGM (the original detection problem); (iii) identify the TGM that generated a given article (similar to Tay et al.,~\shortcite{tay_arxiv20}). For the original detection problem, the authors find that the text generated by the GPT-2 model to be hard to detect among several TGMs (see Appendix for the list of studied TGMs).

\subsection{Zero-shot classifier}
\label{sec:zero_shot}
In the zero-shot classification setting, a pretrained TGM (for example, GPT-2, GROVER) is employed to detect generations from itself or similar models. The detector does not require supervised detection examples for further training (i.e., fine-tuning). 

\noindent\textbf{\underline{Total log probability}}: Solaiman et al.,~\shortcite{solaiman_arxiv19} present a baseline that uses TGM to evaluate total log probability, and thresholds based on this probability to make the prediction. For instance, text is predicted as machine generated if the overall likelihood of the text according to the GPT-2 model is closer to the mean likelihood over all machine generated
texts than to the mean likelihood of human written
texts. However, they find that this classifier performs poorly compared to the previously discussed 
logistic regression based classifier (\textsection\ref{sec:classi_from_scratch}). 

\noindent\textbf{\underline{Giant Language model Test Room (GLTR) tool}}: The GLTR tool~\cite{gehrmann_acl19} proposes a suite of baseline statistical methods that can highlight the distributional differences in text generated by GPT-2 model and human written text. Specifically, GLTR enables the study of a piece of text by visualizing per-token model probability, per-token rank in the predicted next token distribution, and entropy of the predicted next token distribution. Based on these visualizations, the tool clearly shows that TGMs over-generate from a limited subset of the true distribution of natural language. Indeed, rare word usage in text generated by GPT-2 model is markedly less compared to the human written text. The tool lets humans (including non-experts) to study a piece of text, but might be less effective in future once TGMs start generating text that lacks statistical anomalies.

\subsection{Fine-tuning NLM}
\label{sec:finenlm}
In this setup, a pretrained language model (e.g., BERT, RoBERTa~\cite{liu_roberta19}) is fine-tuned to detect text generated from itself or similar models. Unlike the zero-shot classification setup, the detector does require supervised detection examples for further training. 

\noindent\textbf{\underline{GROVER detector}}: Zellers et al.,~\shortcite{zellers_neurips19} propose a detector based on a linear classifier on top of GROVER model, which outperforms existing detectors (fastText~\cite{bojanowski-etal-2017-enriching} and BERT~\cite{devlin_naacl19}) and thereby conclude that the best models for generating neural disinformation are also the best at detecting their own generations. This result suggests the need to make generators such as GROVER and GPT-2 publicly available.~\footnote{The public release of TGM is a complex issue that warrants interdisciplinary considerations, including from policy and security groups. The authors of the GPT-2 model~\cite{gpt_radford18} initially kept their largest model private due to concerns about the potential for misuse. They released their largest model, eight months after publication of the article.} Nevertheless, the authors do not experiment with BERT model to observe similar pattern that the BERT model also excels in detecting the text written by itself as the BERT detector and the BERT generator possess similar inductive bias. Uchendu et al.,~\shortcite{uchendu_emnlp20} show that the off-the-shelf GROVER detector does not perform well in detecting text generated by TGMs other than the original GROVER model.

\noindent\textbf{\underline{RoBERTa detector}}: Solaiman et al.,~\shortcite{solaiman_arxiv19} experiment with fine-tuning the RoBERTa language model for the detection task and establishes the state-of-the-art performance in identifying the web pages generated by the largest GPT-2 model with $\sim$95\%  accuracy. 
The RoBERTa detector trained on top-p examples transfers well to examples from all the other decoding methods (pure and top-k). Regardless of the detector model's capacity, the detector performs well when trained on examples from the larger GPT-2 model and transfers well to examples generated by a smaller GPT-2 model. On the other hand, training on smaller GPT-2 model's outputs results in poor performance in classifying the larger GPT-2 model's outputs. The most interesting finding of this work is that fine-tuning using the RoBERTa model achieves higher accuracy than fine-tuning a GPT-2 model with equivalent capacity. \laks{This result might be due to the superior quality of the bidirectional representations inherent in the masked language modeling objective employed by the RoBERTa language model compared to the GPT-2 language model, which is limited by learning only unidirectional representation (left to right).}  This finding contradicts that of the GROVER work~\cite{zellers_neurips19}, where the authors conclude that the best models for detecting neural disinformation from a TGM is the TGM itself. Recently, Fagni et al.,~\shortcite{fagni2020tweepfake} show that the RoBERTa detector establishes the state-of-the-art performance in spotting machine generated tweets from human written tweets accurately, outperforming both traditional ML models (e.g., bag-of-words) and complex neural network models (e.g., RNN, CNN) by a large margin. This interesting result indicates that the RoBERTa detector can generalize to publication sources unseen during its pretraining such as Twitter. The RoBERTa detector also outperforms existing detectors in spotting news articles generated by several TGMs~\cite{uchendu_emnlp20} and product reviews generated by the GPT-2 model fine-tuned on Amazon product reviews~\cite{adelani_aina20}.    

\subsection{Human-machine collaboration}
\label{sec:human_machine}
Apart from building a statistical model to detect online disinformation, one can build a system that can leverage human visual interpretation skills and common sense knowledge. 

\noindent\textbf{\underline{Differences in human and machine detector}}: Ippolito et al.,~\shortcite{ippolito_arxiv19} study the differences in the ability of humans and automated detectors to identify text generated by TGM. The authors observe: (i) human raters are good at noticing contradictions or semantic errors (e.g., incoherence) in text generated by TGM, which the automatic detectors are weak at, due to lack of deep semantic understanding; (ii) automatic detectors are good when text generated by TGM contains over-representation of high-likelihood words (caveat of top-k sampling as discussed in \textsection\ref{sec:related_gen_text_tgm}), whereas the human raters are not good. Overall, automatic detectors are significantly better than human raters, but generalize poorly to text generated by unseen decoding methods. 

\noindent\textbf{\underline{Supporting untrained humans}}: As seen before, the GLTR tool~\cite{gehrmann_acl19} can aid humans by visualizing the properties of text such as unexpected and out-of-context words. The main advantage of GLTR is that it can facilitate untrained humans to accurately detect synthetic text (from 54\% to 72\% in terms of accuracy). However, GLTR flags \eat{text as}\laks{machine generated easily} but \laks{it is} hard to be confident that the text is \textit{not} machine generated. This result suggests the need for human-machine collaboration to solve the detection task~\cite{solaiman_arxiv19}.\note[Laks]{Is this disinformation problem OR the problem of distinguishing TGM text from human generated one?}\gan{I changed it to detection task.}

\noindent\textbf{\underline{Real or Fake Text (RoFT) tool}}: The RoFT tool~\cite{dugan2020roft} focuses on evaluating human detection of text generated by TGM by asking humans to detect the sentence boundary at which the text transitions from human written text to machine generated text. The main assumption is that TGM successfully fools the human if the guess from the human is far from the true sentence boundary. Current TGMs can fool humans by one or two sentences. The core advantages of the RoFT tool include its engaging annotation interface, collection of user's explanation for their guess in free form text, and potential to scale to different textual domains as well as different TGM modeling choices. The main limitation of the tool is that the text shown to the humans can be rife with human generated sentences, and hence does not reflect an organic generation from a TGM.

\section{Issues with the state-of-the-art detector}
\label{sec:roberta_issues}
In this section, we discuss open issues in the state-of-the-art detector based on the RoBERTa model, which has been shown to excel in detecting text generated by TGM based on news articles, product reviews, tweets, and web pages (see \textsection\ref{sec:finenlm}).~\footnote{Concurrent with our work, Zhong et al.,~\shortcite{zhong2020neural} propose a detector that leverages factual and coherence structure underlying the text, which outperforms the RoBERTa detector in spotting machine generated text based on news articles and web pages. We also acknowledge that detectors fine-tuned on the state-of-the-art NLMs such as T5~\cite{t5_raffel20}, ELECTRA~\cite{Clark2020ELECTRA} might most likely outperform the RoBERTa detector in general.} We focus on the task of detecting text generated by the GPT-2 model from human written Amazon product reviews, \laks{a challenging task given the shortness of reviews.} We employ the RoBERTa detector on the publicly available dataset, containing generations from the GPT-2 model ($1542M$ parameters) based on pure, top-$k$ and top-$p$ sampling along with human written reviews (see Appendix for dataset details).
 In Figure~\ref{fig:det_acc}, we plot the accuracy of the detector w.r.t. number of training examples per class, averaged over ten random initializations to control for initialization effects. 
We observe that the RoBERTa detector \eat{seems \textbf{data-inefficient}, that is, the detector}needs several thousands of examples to reach high accuracy. Specifically, it has an impractical requirement of $200K$, $15K$ and $50K$ training examples for performing at $90\%$ accuracy on identifying pure, top-k and top-p examples respectively.~\footnote{Given that attackers can create synthetic text at scale using TGMs, 90\% detection accuracy might not be a high accuracy.} Given that creation of large datasets for the detection task is hard~\cite{zellers_neurips19}, it is important to investigate whether the \textit{data-efficiency} of the RoBERTa detector can be significantly improved. 
\eat{ 
\note[Laks]{Roberta is data-inefficient compared to what? Making an absolute statement w/o a standard or reference does't make sense to me. Same comment about the "impractical requirement" claim below. What is your "standard"?}\gan{I don't have a standard. ~\cite{zellers_neurips19} mentioned it's hard to collect such a dataset from scratch, as number of fake news articles are extremely less and it's hard to pinpoint to a single generated model. I've changed it from absolute statement. Is it ok?}}
\note[Laks]{Rewrote slightly. See if it's OK.}
\begin{figure}[ht]
\centering
\includegraphics[width=3.4in, height=1.6in]{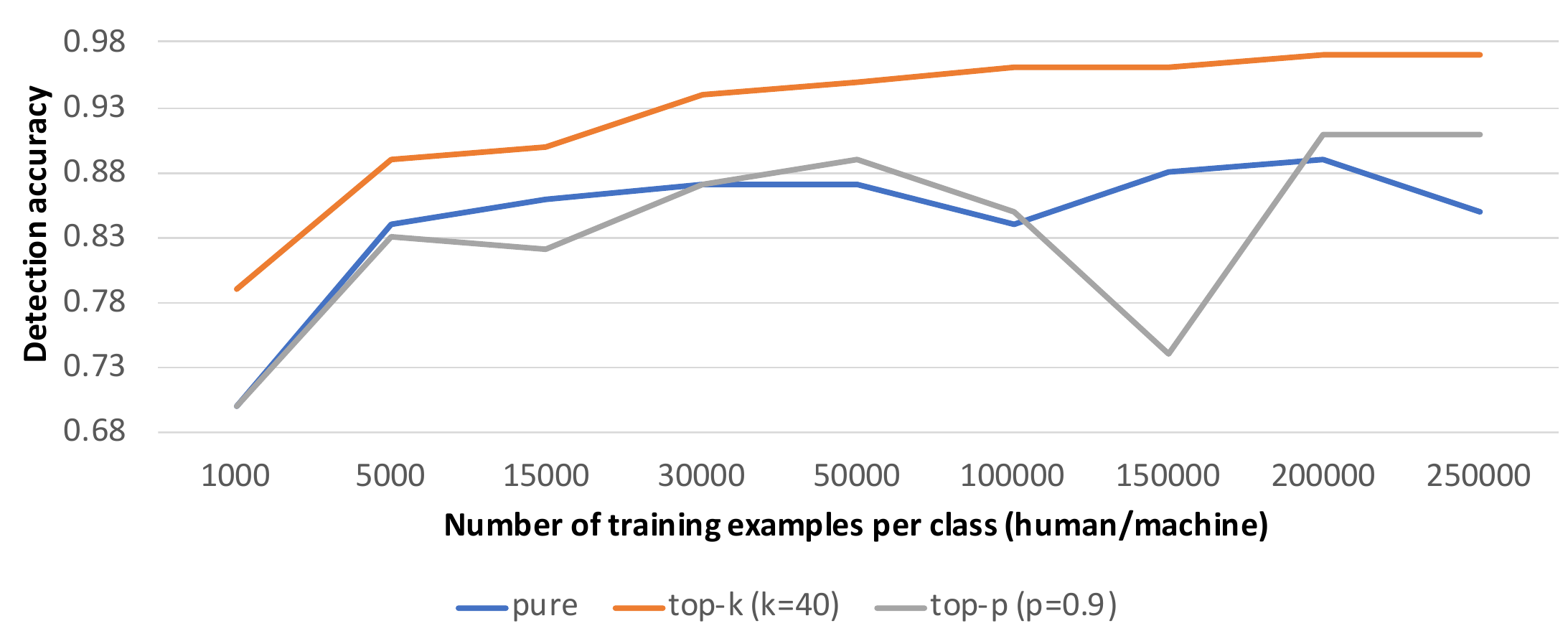}
\caption{Detection accuracy of the RoBERTa detector w.r.t. number of training examples per class, averaged over ten random initializations.}
\label{fig:det_acc}
\end{figure}

We manually inspect 100 randomly picked false positives (machine generated product review incorrectly predicted as human written product review) of the RoBERTa detector trained on $15K$ examples each from top-$p$ generations and from human written reviews.\footnote{As seen in~\textsection\ref{sec:related_gen_text_tgm} and ~\textsection\ref{sec:det}, top-p sampling produces good quality text that reasonably matches the style of human writing and is also harder to detect for humans. We leave the study of false negatives for future. Our annotation of 100 false positives can be accessed at:~\url{https://github.com/UBC-NLP/coling2020_machine_generated_text}.}
\eat{ 
\note[Laks]{I don't understand this parenthetical remark.}\gan{Detecting top-p samples is the most difficult task compared to top-k and pure detection.} 
}Below, we list down the error categories that we have identified and provide at least one example for each error category. 

\noindent\textbf{\underline{Fluency}}: Among the false positive reviews, we find $73$ reviews to be very \textit{fluent} and can confuse even humans (1). 

\begin{tabular}{p{0.3in}p{5.5in}}
(1) & I loved this film. I can't really explain why, but when I first saw it it struck me as bizarre, almost oddball, but I quickly got over that and remembered that I love oddball films. This was an early 80's film. A great film to see on a gloomy rainy evening. This film is suspenseful and full of weirdness. Add this to your collection.
\end{tabular}


\noindent\textbf{\underline{Shortness}}: Out of these $73$ identified fluent reviews, $27$ reviews are very \textit{short}, with a median of 24 words. \laks{We give two examples below:} 
\eat{ 
\note[Laks]{Why is the median an interval?}\gan{Those (2), (3) are example ids.}
} 

\begin{tabular}{p{0.3in}p{5.5in}}
(2) & love it. best sweeper. \\
(3) & My favorite combo. Always works and usually cools my system to boot. So glad I got these instead of other brands.
\end{tabular}


\noindent\textbf{\underline{Factuality}}: We find $10$ false positive reviews to contain \textit{factual} errors. 

\begin{tabular}{p{0.3in}p{5.5in}}
(4) & That movie got the stars and represents the best of this collection but there's better made Creature Movies as well including a 1960's remake of `Dracula' with Kirk Douglas and \underline{Harrison Ford}. \\
(5) & Just love \underline{Ben Affleck}! He won't be missed in another very good movie. Worth watching especially if you like Ben!
\end{tabular}

Review (4) on product `Universal Studios Classic Monster Collection'  contains the incorrect fact that Harrison Ford  acted in `Dracula' movie, and another review (5) on `Runaway Jury' movie  contains the incorrect fact that Ben Affleck  acted in that movie.

\noindent\textbf{\underline{Spurious entities}}: In $4$ false positive reviews, we find that the review contains \textit{novel entities} unrelated to the domain of the product. For example, review (6) on `Junkfood' musical product contains novel entity, `grisberg', 
which is not associated with music domain. 
\note[Laks]{What is the technical difference between containing novel entities and having incorrect facts? Can't we argue that Harrison Ford is a "novel" entity w.r.t. the `Dracula' movie?}\gan{It's more like non-existent entities. Stevie is probably a singer, but there seems to be no entity "grisberg" in music domain (based on my google search)}

\begin{tabular}{p{0.3in}p{5.5in}}
(6) & another classic by \underline{grisberg}, i love stevie she was one of the greatest r\&b singers I know darwin halstead ment her so be a big fan please do yo self a favor and buy this dvd, its nice and it absolutly amazing this woman has a very yorfelt approch to r\&b music
\end{tabular}

\noindent\textbf{\underline{Contradiction}}:
We find one review (7) containing \textit{contradictions}, \laks{where the subject (husband) is \eat{contradictorily mentioned} claimed to be not a big fan of a product but also as  \textit{loving} the same product.} 

\begin{tabular}{p{0.3in}p{5.5in}}
(7) & My husband likes his coffee black so he \underline{loves} flavored coffee but is \underline{not a big fan} of flavored coffee. ...
\end{tabular}

\noindent\textbf{\underline{Repetition}}: In two false positive reviews, the facts undergo \textit{repetition}. 

\begin{tabular}{p{0.3in}p{5.5in}}
(8) & Great movie, although took a while to see at first it held my interest and kept me interested, plus i thought it was \underline{extremly good}. also it was \underline{very good}.
\end{tabular}


\noindent\textbf{\underline{Common sense reasoning}}: We find one false positive review that describes an improbable event, that is, violates \textit{common sense reasoning}. 


\begin{tabular}{p{0.3in}p{5.5in}}
(9) & ... I received both \underline{amazon} Prime and a \underline{Walmart's} for delivery and they both came on time. I love it and highly recommend it!
\end{tabular}

The review (9) on a specific audio player product mentions that the user received the same product from two e-commerce companies simultaneously, which is most likely an improbable event.

\noindent\textbf{\underline{Typos and grammatical errors}}: There are $7$ false positive reviews that possess \textit{typographical} and \textit{grammatical} errors (10) and (11). We note that such errors (especially spelling errors) are not unusual in online reviews, \laks{including those by humans}. 
\note[Laks]{(11) seems like a combination of typo plus grammatical errror. Even after changing don to done, it doesn't read right.}\gan{Changed incompleteness to grammatical errors.}

\begin{tabular}{p{0.3in}p{5.5in}}
(10) & Once they are on they aren't wrinkled or lose \underline{they} shape. \\
(11) & Had to unplug thing to get the hard drive to work. Would rather have \underline{don} batteries in the olden days..
\end{tabular}

\noindent\textbf{\underline{Incoherence}}:
There are $3$ false positive reviews that seem \textit{incoherent}. The movie review (12) switches the focus of the discourse between actors (Sophia and Duchovny) and story line in an incoherent fashion, which violates the theory of centering in discourse analysis~\cite{grosz_acl95,gehrmann_acl19}.

\begin{tabular}{p{0.3in}p{5.5in}}
(12) & ...  Sophia Loren plays `Marion' a `showgirl' that is picked on by the establishment for her wild style. ... Duchovny's character is also `On the line' in the business world. ... The storyline is so intriguing and unpredictable. ... Sophia Loren's acting is just awesome and her wardrobe is just perfect! If you love sex and nud**y, you will be greatly pleased.
\end{tabular}

\section{Future Research Directions}
\label{sec:future}
In this section, we  discuss a set of  future research directions, which can help in building useful detectors.

\subsection{Leveraging auxiliary signals}

\note[mam]{``helpfulness" seems more like a ``label", rather than ``meta-information". Also, ``description" of a product doesn't come along, for me, as ``meta-information", but I could live with that one. But, is there a `better' term than `meta-information' that we can employ here? ``Meta-data" seems something people would be used to more.}\gan{Idk. I can change it to metadata.}\note[Laks]{Following up on this comment, I agree that meta-info is not a standard term. OTOH, meta-data is normally more structured, sth you wouldn't have to mine. Need a good term here. What about "auxiliary signals"? It's not the best but perhaps you can improve on it.}\note[mam]{I like "auxiliary signals".}
Existing detectors do not exploit auxiliary signals about the textual source.~\footnote{Concurrent with our work, Tan et al.,~\shortcite{tan2020detecting} propose a detector that spots machine generated news articles by utilizing news body, images, and captions associated with the news articles.} For example, the RoBERTa detector studied in \textsection\ref{sec:roberta_issues} 
ignores the auxiliary signals about the review (e.g., helpfulness) and the product (e.g., description). 
Such auxiliary signals can be complementary to linguistic signals from the textual source for the detection task~\cite{hovy-2016-enemy,solaiman_arxiv19}. Given the rapidly evolving research in building intelligent TGMs that narrows the gap between machine and human distribution of natural language text, auxiliary signals could play a crucial role in mitigating the threats posed by TGMs.      

\subsection{Assessing veracity of the text}
Existing detectors have an assumption that the fake text is determined by the source (e.g., TGM) that generated the text. 
This assumption does not hold true in two practical scenarios:
(i) real text 
auto-generated in a  process similar to that of fake text, and (ii) adversaries 
creating fake text by modifying articles originating from legitimate human sources. 
Schuster et al.,~\shortcite{schuster_cl20} show that existing detectors perform poorly in these two scenarios as they rely too much on distributional features, which cannot help in distinguishing texts from similar sources.
Hence, we call for more research on detectors that assess the veracity of machine generated text by consulting external sources, like knowledge bases~\cite{thorne-vlachos-2018-automated} and diffusion network~\cite{Vosoughi1146}, \laks{instead of relying only on the source.} 

\subsection{Building generalizable detectors}
Existing detectors exhibit poor cross-domain accuracy, that is, they are not generalizable to \laks{different} publication formats (Wikipedia, books, news sources)~\cite{bakhtin_arxiv19}.\note[Laks]{Changed wide range of to different since we give just 2 examples. Change back if you feel strongly about it.}\gan{It seems fine. I can add news as well.} Beyond publication formats and topics (e.g., politics, sports), the detector should also transfer to unseen TGM settings such as model architecture, different decoding methods (e.g., top-$k$, top-$p$), model size, different prefix lengths, and training data~\cite{bakhtin_arxiv20,uchendu_emnlp20}.

\subsection{Building interpretable detectors}
We discussed the importance of human raters pairing up with automatic detectors in  \textsection\ref{sec:human_machine}. A viable way for this collaboration is to make the decisions taken by the automatic detector interpretable (such as in GLTR) so that human raters can logically group (e.g., contradictions) the model decisions and humans can ``accept'', ``modify'', or ``reject'' these decisions. \laks{This calls for} more research in building detectors that can provide explanations for its decisions, which are understandable to humans.

\subsection{Building detectors robust to adversarial attacks}
Existing detectors are brittle, i.e., the detector decisions can vary significantly for even small changes in the text input. For example, Wolff~\shortcite{wolff2020attacking} shows  that the RoBERTa detector can be attacked using simple schemes such as replacing characters with homoglyphs and misspelling some words. These two attacks reduce the detector's recall in text generated by TGM from 97.44\% to 0.26\% and 22.68\% respectively.\note[Laks]{Checking that you really meant 0.26\% rather than 26\%.}\gan{It's 0.26\%.} Therefore, it is important to study various adversarial attacks ranging from simple attacks (e.g., misspellings) to advanced attacks (e.g., universal attacks~\cite{wallace-etal-2019-universal}) and  create adversarial examples with an aim to characterize the vulnerabilities of the detector as well as to \laks{make the detector robust against various attacks.} 

\section{Conclusion}
\label{sec:conclusion}
Detectors able to tease apart machine generated text from human written text can play a vital role in mitigating misuse of TGMs such as in automatic creation of fake news and fake product reviews. Our categorization of existing detectors and related issues into classifiers trained from scratch, zero-shot classifiers, fine-tuning NLMs, and human-machine collaboration can help readers contextualize each detector w.r.t the fast-growing literature. We also hope that our computationally and linguistically motivated error analysis of the state-of-the-art detector can bring readers up to speed on many existing challenges in building useful detectors. Our rich and diverse set of research directions also have the potential to guide future work in this exciting area.
\note[Laks]{Can be strengthened by discussing the taxonomy based on methodology, error analysis, etc. }

\section*{Acknowledgements}
We thank Ramya Rao Basava and Peter Sullivan for helpful discussions in the initial stage of the project. 
We gratefully acknowledge support from the Natural Sciences and Engineering Research Council of Canada, Compute Canada (\url{https://www.computecanada.ca}), and UBC ARC–Sockeye (\url{https://doi.org/10.14288/SOCKEYE}).

\bibliographystyle{coling}
\bibliography{main}

\section*{Appendix - Existing detection datasets}
\label{sec:related_det_benchmarks}

\begin{table*}[htb]
\footnotesize
\begin{center}
\begin{tabular}{p{0.5in}|p{2.2in}|l|l|l|l|l|l}
            \multicolumn{2}{c|}{Split} & \multicolumn{2}{c|}{Train Set} & \multicolumn{2}{c|}{Validation Set} & \multicolumn{2}{c}{Test Set} \\ \hline 
human & machine & human & machine & human & machine & human & machine \\ \hline
WebText & Article from GPT-2 with pure sampling & 250K & 250K & 5K & 5K & 5K & 5K \\ 
WebText & Article from GPT-2 with top-$k$ sampling & 250K & 250K & 5K & 5K & 5K & 5K \\ 
WebText & Article from GPT-2 with top-$p$ sampling & 250K & 250K & 5K & 5K & 5K & 5K \\ \hline
Amazon Review & Review from GPT-2 finetuned on Amazon Reviews with pure sampling & 250K & 250K & 5K & 5K & 5K & 5K \\ 
Amazon Review & Review from GPT-2 finetuned on Amazon Reviews with top-$k$ sampling & 250K & 250K & 5K & 5K & 5K & 5K \\ 
Amazon Review & Review from GPT-2 finetuned on Amazon Reviews with top-$p$ sampling & 250K & 250K & 5K & 5K & 5K & 5K \\ \hline
RealText & News articles from GROVER with top-$p$ sampling & 5K & 5K & 2K & 1K & 8K & 4K \\ \hline
Tweet & Tweet by markov chain or RNN or GPT-2 or misc. & 10358 & 10354 & 1150 & 1152 & 1278 & 1280 \\ \hline
- & Article by GPT-3 with top-$p$ sampling & - & - & - & - & - & 2008 \\ \hline
News & Article by several TGMs & 1066 & 8528 & - & - & - & - \\ \hline
\end{tabular}
\caption{Dataset statistics for existing detection datasets.}
\label{tab:datastats}
\end{center}
\end{table*}

Here, we will discuss existing detection datasets in the literature. Table~\ref{tab:datastats} displays the statistics for all the detection datasets, which we will introduce now.  

\noindent\textbf{\underline{WebText vs. GPT-2}}: The Generative Pre-trained Transformer 2 (GPT-2) model~\cite{radford2019language} is originally trained on WebText, a collection of online articles, sourced from high quality outbound links from Reddit. Solaiman et al.,~\shortcite{solaiman_arxiv19} provide articles generated by GPT-2 based on pure, top-$k$, and top-$p$ sampling methods.~\footnote{\url{https://github.com/openai/gpt-2-output-dataset}} Since online articles can come from different domains, this WebText dataset lets us study the generalizability of the detector with respect to the domain of the text.

\noindent\textbf{\underline{Amazon Product Reviews vs. GPT-2}}: Solaiman et al.,~\shortcite{solaiman_arxiv19} finetuned the GPT-2 model on Amazon product reviews~\cite{amazon_review} to make GPT-2 generate a product review that reasonably matches the style of Amazon product review. Similar to the WebText dataset, the authors provide reviews generated by GPT-2 based on pure, top-$k$, and top-$p$ sampling methods. This review dataset makes the detection task challenging due to lack of context (reviews are short with a median of 115, 141 words for human and top-$p$ machine reviews respectively). 

\noindent\textbf{\underline{RealNews vs. GROVER}}: The Generating aRticles by Only Viewing mEtadata Records (GROVER) model~\cite{zellers_neurips19} is trained on RealNews, a collection of news articles from Common Crawl. The authors of the GROVER model provide a subset of news articles (not part of the training set of GROVER model) and news articles generated by GROVER model with top-$p$ sampling.~\footnote{\url{https://github.com/rowanz/grover/tree/master/generation_examples}} 

\noindent\textbf{\underline{Tweets vs. Misc.}}: Social media platforms like Twitter has several bot user accounts, whose entire timeline is composed of tweets produced by models such as markov chain, RNN, LSTM~\cite{lstm}, GPT-2, and several miscellaneous (unknown) models. Fagni et al.,~\shortcite{fagni2020tweepfake} provide a collection of tweets from manually identified bot accounts and a collection of tweets from the humans imitated by the bot accounts.~\footnote{\url{https://www.kaggle.com/mtesconi/twitter-deep-fake-text}} This tweet dataset is challenging as the tweets are extremely short (median of 14, 16 words for human and machine tweets respectively). Unlike other datasets, this tweet dataset contains real machine generated texts posted in Twitter, which can directly measure the real world utility of the detector. Since these machine generated tweets encompass generations from different TGM models such as markov chain, LSTM, GPT-2 and miscellaneous models, this tweet dataset lets us study the generalizability of the detector with respect to the TGM that produced the text.

\noindent\textbf{\underline{GPT-3}}: The GPT-3~\cite{brown2020language} model is trained on WebText, Wikipedia, Books and Common Crawl. The authors of the GPT-3 model provide generations of GPT-3 with top-$p$ sampling.~\footnote{\url{https://github.com/openai/gpt-3/blob/master/175b_samples.jsonl}} Similar to the WebText dataset, this GPT-3 dataset lets us study the generalizability of the detector with respect to the domain of the text.

\noindent\textbf{\underline{Politics-News vs. Misc.}}: Uchendu et al.,~\shortcite{uchendu_emnlp20} provide human written news articles related to politics category. Utilizing the title of the human written news article as prompt, the authors generate corresponding machine generated article from eight TGMs, which includes CTRL~\cite{keskar_arxiv19}, GPT-1~\cite{gpt_radford18}, GPT-2~\cite{radford2019language}, GROVER~\cite{zellers_neurips19}, XLM~\cite{xlm}, XLNet~\cite{xlnet}, PPLM~\cite{dathathri_iclr20}, and FAIR~\cite{ng-etal-2019-facebook}.~\footnote{\url{https://github.com/AdaUchendu/Authorship-Attribution-for-Neural-Text-Generation}} Similar to the tweets dataset, this news dataset lets us study the generalizability of the detector with respect to the TGM that produced the text.

\end{document}